\documentclass[english]{article}
\usepackage{lmodern}
\usepackage[T1]{fontenc}
\usepackage[latin9]{luainputenc}
\usepackage{amsmath}
\usepackage{graphicx}
\usepackage[numbers]{natbib}

\makeatletter

\providecommand{\tabularnewline}{\\}
\usepackage[final]{nips_2017}

\usepackage{hyperref}       % hyperlinks
\usepackage{url}            % simple URL typesetting
\usepackage{booktabs}       % professional-quality tables
\usepackage{amsfonts}       % blackboard math symbols
\usepackage{nicefrac}       % compact symbols for 1/2, etc.
\usepackage{microtype}      % microtypography
\usepackage{times}
\usepackage{helvet}
\usepackage{courier}

\usepackage{amsbsy}
\usepackage{algorithmic}
\usepackage[english]{babel}

\usepackage{microtype}
\usepackage{flushend}

\makeatother

\usepackage{babel}
\begin{document}

\title{Finding Algebraic Structure of Care in Time: A Deep Learning Approach}

\author{Phuoc Nguyen, Truyen Tran and Svetha Venkatesh\\
Centre for Pattern Recognition and Data Analytics\\
Deakin University, Geelong, Australia\\
 \{phuoc.nguyen,truyen.tran,svetha.venkatesh\}@deakin.edu.au}

\maketitle
\global\long\def\xb{\boldsymbol{x}}
\global\long\def\yb{\boldsymbol{y}}
\global\long\def\hb{\boldsymbol{h}}
\global\long\def\Xcal{\mathcal{X}}
\global\long\def\Ucal{\mathcal{U}}
\global\long\def\Vcal{\mathcal{V}}
\global\long\def\Real{\mathbb{R}}
\global\long\def\mat{\text{mat}}
\global\long\def\thetab{\boldsymbol{\theta}}
\global\long\def\Wb{\boldsymbol{W}}
\global\long\def\bb{\boldsymbol{b}}
\global\long\def\cb{\boldsymbol{c}}
\global\long\def\fb{\boldsymbol{f}}
\global\long\def\gb{\boldsymbol{g}}
\global\long\def\ib{\boldsymbol{i}}
\global\long\def\ob{\boldsymbol{o}}
\global\long\def\ub{\boldsymbol{u}}
\global\long\def\vb{\boldsymbol{v}}
\global\long\def\db{\boldsymbol{d}}
\global\long\def\pb{\boldsymbol{p}}
\global\long\def\eb{\boldsymbol{e}}
\global\long\def\Db{\boldsymbol{D}}
\global\long\def\Ib{\boldsymbol{I}}
\global\long\def\Hb{\boldsymbol{H}}
\global\long\def\qb{\boldsymbol{q}}

\begin{abstract}
Understanding the latent processes from Electronic Medical Records
could be a game changer in modern healthcare. However, the processes
are complex due to the interaction between at least three dynamic
components: the illness, the care and the recording practice. Existing
methods are inadequate in capturing the dynamic structure of care.
We propose an end-to-end model that reads medical record and predicts
future risk. The model adopts the algebraic view in that discrete
medical objects are embedded into continuous vectors lying in the
same space. The bag of disease and comorbidities recorded at each
hospital visit are modeled as function of sets. The same holds for
the bag of treatments. The interaction between diseases and treatments
at a visit is modeled as the residual of the diseases minus the treatments.
Finally, the health trajectory, which is a sequence of visits, is
modeled using a recurrent neural network. We report preliminary results
on chronic diseases \textendash{} diabetes and mental health \textendash{}
for predicting unplanned readmission.

\end{abstract}

\section{Introduction}

An ultimate goal of AI in healthcare is to reason about the past (historical
illness), present (diagnosis) and future (prognosis). The learning
path through recorded medical data necessitates the modeling of the
dynamic interaction between the three processes: the illness, the
care and the data recording \cite{pham2017predicting}. For this paper,
we assume that a hospital visit at a time manifests through a set
of discrete diseases and treatments. A healthcare trajectory is therefore
a sequence of time-stamped records.

Here we adopt the notion of reasoning as ``algebraically manipulating
previously acquired knowledge in order to answer a new question''
\cite{bottou2014machine}. For that we learn to embed discrete medical
objects into continuous vectors, which lend themselves to a wide host
of powerful algebraic and statistical operators \cite{choi2016multi,tran2015learning}.
For example, with diseases represented as vectors, computing disease-disease
correlation is simply a cosine similarity between the two corresponding
vectors. Illness \textendash{} recorded as a bag of discrete diseases
\textendash{} can then be a function of set of vectors. The same holds
for care. Importantly, if diseases, treatments (or even doctors) are
embedded in the same space then recommendation of treatments (or doctors)
for a given disease will be as simple as finding the nearest vectors.

The algebraic view makes it easily to adapt powerful tools from the
recent deep learning advances \cite{lecun2015deep} for healthcare.
In particular, we can build end-to-end models for risk prediction
without manual feature engineering \cite{cheng2016risk,nguyen2016deepr,pham2017predicting,ravi2017deep}.
As the models are fully differentiable, credit assignment to distant
risk factors can be carried out \cite{nguyen2016deepr}, making the
models more transparent than commonly thought. 

One important aspect still remains, however, that is the dynamic interaction
between illness and care. Although care is supposed to lessen the
illness, it is often designed through highly controlled trials where
one treatment is targeted at one disease, on a specific cohort, at
a specific time. Much less is known for the effect of multiple treatments
on multiple diseases, in general hospitalized patients, \emph{over
time}. A recent model known as DeepCare \cite{pham2017predicting}
partly addresses this problem by considering the moderation effect
of treatments on illness state transition \emph{between visits}, but
not multi-disease\textendash multi-treatment interaction \emph{within
visits}.

This paper reports preliminary results on an investigation into finding
the algebraic structure of care hidden in the Electronic Medical Records.
The task is to predict future risk such as unplanned readmission or
death at discharge. We focus on chronic diseases (diabetes and mental
health) as they are highly complex with multiple causes, often associated
with multiple cormobidities, and the treatments are not always effective.

\section{Methods\label{sec:Methods}}

We consider the problem of modeling Electronic Medical Records (EMR)
for predicting future risk at the time of hospital discharge. Each
medical record is a sequence of hospital visits by a patient. For
simplicity, we consider a visit as consisting of diseases and treatments.
While we might expect that the disease subset together with the treatment
subset reflect the illness state at the time of discharge, it is not
necessarily the case. This is because of several reasons. First, the
coding of those diseases and treatments is often optimized for billing
purposes, not all diseases are included. Second, errors do occur sometimes.
And third, the treatments usually take time to get the full intended
effect.

For this reason, we include historical visits to assess the current
state as well as to predict future risk. An efficient way is to model
the visit sequences as a Recurrent Neural Network (RNN) \cite{elman1990finding}.
In this paper, we choose Long Short-Term Memory (LSTM) since it can
remember distant events \cite{hochreiter1997long}.

\subsection{Visit Representation}

A visit consists of two variable-size bags of discrete elements: a
bag of diseases and a bag of treatments. Among older cohorts, non-singleton
bags are prevalent, reflecting the comorbidity picture of modern healthcare.
As a result, treatments must be carefully administered to work with,
or at least not to cancel out, each other. This also calls for a sensible
way to model the complexity of \emph{multi-disease\textendash multi-treatment
interaction}. Most existing biostatistics methods, however, are designed
for simplified treatment effect against just one condition.

We use vector representation of diseases and treatments, following
the recent practice in NLP (e.g., see \cite{collobert2011natural}).
Let $\eb_{d}$ be the vector representation of disease $d$, $\eb_{p}$
the representation of the treatment $p$, and the vectors are embedded
in a common space. The bags of diseases and bags of treatments are
also represented as vectors in the same space. The representation
of a bag is computed using a differentiable set function $f_{e}(S)$
that receives a bag of vectors $S$ and returns another vector of
the same dimensions. We use the following set function:

\begin{align}
f_{e}(S) & \leftarrow\frac{\bar{\eb}_{S}}{\epsilon+\left\Vert \bar{\eb}_{S}\right\Vert }\quad\text{where}\quad\bar{\eb}_{S}=\max\text{(\ensuremath{\mathbf{0}},\ensuremath{\sum_{i\in S}\eb_{i}})}\label{eq:set-func}
\end{align}
where $\epsilon>0$ is a smoothing factor. This is essentially a linear
rectifier \cite{nair2010rectified} of the sum, approximately normalized
to unit vector. The factor $\epsilon$ lets $\left\Vert f_{e}(S)\right\Vert \rightarrow0$
when$\left\Vert \bar{\eb}_{S}\right\Vert \rightarrow0$, but $\left\Vert f_{e}(S)\right\Vert \rightarrow1$
when$\left\Vert \bar{\eb}_{S}\right\Vert \gg0$.

Denote by $D_{t}$ the bag of diseases and $I_{t}$ the bag of treatments
recorded for the visit at time $t$. Let $\db_{t}=f_{e}\left(D_{t}\right)$
be the representation of the disease bag, and $\pb_{t}=f_{e}\left(I_{t}\right)$
the representation of the treatment bag. We compute the visit vector
as:

\begin{equation}
\vb_{t}=\rho\left(\Delta\right)\quad\text{where}\quad\Delta=\db_{t}-\pb_{t}\label{eq:disease-treatment-effect}
\end{equation}
where $\rho$ is an element-wise transformation. The difference $\Delta$
reflects the intuition that treatments are supposed to lessen the
illness. We found $\rho\left(\Delta\right)=\left(1+\Delta\right)^{2}$
works well, suggesting that the disease-treatment interaction is highly
nonlinear, and thus warranting a deeper investigation.

\subsection{Illness Memory with LSTM}

Given a sequence of input vectors, one per visit, the LSTM reads an
input $\vb_{t}$ at a time and estimates the illness state $\hb_{t}$.
To connect to the past, LSTM maintains an internal short-term memory
$\cb_{t}$, which is updated after seeing the input. Let $\tilde{\cb}_{t}$
be the new candidate memory update after seeing $\vb_{t}$, the memory
is updated over time as $\cb_{t}=\fb_{t}\ast\cb_{t-1}+\ib_{t}\ast\tilde{\cb}_{t}$,
where $\fb_{t}\in(\boldsymbol{0},\boldsymbol{1})$ is forget gate
determining how much of past memory to keep; $\ib_{t}\in(\boldsymbol{0},\boldsymbol{1})$
is the input gate controlling the amount of new information to add
into the present memory. The input gate is particularly useful when
some recorded information is irrelevant to the final prediction tasks.

The memory gives rise to the state as $\hb_{t}=\ob_{t}\ast\tanh(\cb_{t})$,
where $\ob_{t}\in(\boldsymbol{0},\boldsymbol{1})$ is the output gate,
determining how much external information can be extracted from the
internal memory. The candidate memory and the three gates are parameterized
functions of $(\vb_{t},\hb_{t-1})$.

With this long short-term memory system in place, information of the
far past is not entirely forgotten, and credit can be assigned to
it. Second, partially recorded information can be integrated to offer
a better picture of current illness.

\subsection{Risk Prediction}

Once the LSTM is specified, its states are pooled for risk prediction
at each discharge, i.e., $\bar{\hb}_{t}=\text{pool}(\hb_{1:t})$.
The pooling function can be as simple as the \texttt{mean()}(i.e.,
$\bar{\hb}_{t}=\frac{1}{t}\sum_{j=1}^{t}\hb_{j}$) or\texttt{ last()}(i.e.,
$\bar{\hb}_{t}=\hb_{t}$). We also tried exponential smoothing, $\bar{\hb}_{t}=\alpha\bar{\hb}_{t-1}+(1-\alpha)\hb_{t}$,
for $\bar{\hb}_{1}=\hb_{1}$ and $\alpha\in[0,1]$. A small $\alpha$
would mean the recent visits have more influence in future risk. 

Finally, a differentiable classifier (e.g., a feedforward neural net)
is placed on top of the pooled state to classify the medical records
(e.g., those in population stratification) or to predict the future
risk (e.g., unplanned readmissions). The loss function is typically
the negative log-likelihood of outcome given the historical observations,
e.g., $-\sum_{t}\log P\left(y_{t}\mid\Db_{1:t},\Ib_{1:t}\right)$.
We emphasize here is the system is end-to-end differentiable, starting
from the disease and treatment lookup table at the bottom to the final
classifier at the top. No feature engineering is needed.

\subsection{Regularizing state transitions}

For chronic diseases, it might be beneficial to regularize the state
transition. We consider adding the regularizers $\frac{\beta}{T}\sum_{t=2}^{T}\left(\left\Vert \hb_{t}\right\Vert _{2}-\left\Vert \hb_{t-1}\right\Vert _{2}\right)^{2}$
to the loss function as suggested in \cite{krueger2015regularizing}.
This asks the amount of information available at each time step, encapsulated
in the norm $\left\Vert \hb_{t}\right\Vert _{2}$, to be stable over
time. This is less aggressive than maintaining state coherence, i.e.,
by minimizing $\frac{\beta}{T}\sum_{t=2}^{T}\left\Vert \hb_{t}-\hb_{t-1}\right\Vert _{2}^{2}$.

\section{Results}

\paragraph{Cohorts}

Data is previously studied in \cite{pham2017predicting}, which consists
of two chronic cohorts: diabetes and mental health. There are over
7000 diabetes patients with a median age of 73 making over 53,000
visits. For mental health cohort, the figures are 6,100, 37 and 52,000
respectively. The number of disease and treatment codes for each cohort
are around 240 and 1100 respectively. Diseases and treatments are
coded using the ICD-10 coding scheme. For diseases, the first two-level
in the ICD-10 tree is used. The data was collected between 2002-2013
from a large regional Australian hospital. Each record contains at
least 2 hospital visits.

\paragraph{Implementation}

Models are implemented in Julia using the Knet.jl package \cite{knet2016mlsys}.
Optimizer is Adam \cite{kingma2014adam} with learning rate of 0.01
and other default parameters. Two baselines are implemented. One is
bag-of-words trained using regularized logistic regression (BoW+LR),
where diseases and treatments are considered as words, and the medical
history as document. No temporal information is modeled. Although
this is a simplistic treatment, prior research has indicated that
BoW works surprisingly well \cite{nguyen2016deepr,pham2017predicting}.
The other is a recent model known as Deepr \cite{nguyen2016deepr},
which is based on convolutional net for sequence classification. Unlike
the BoW, which are unordered, in Deepr words are sequenced by their
temporal order. Words of the same visit are randomly sequenced. However,
the Deepr does not model the temporal transition between illness states.

\paragraph{Visualization}

\begin{figure*}[!t]
\begin{centering}
\begin{tabular}{cc}
\includegraphics[width=0.4\textwidth]{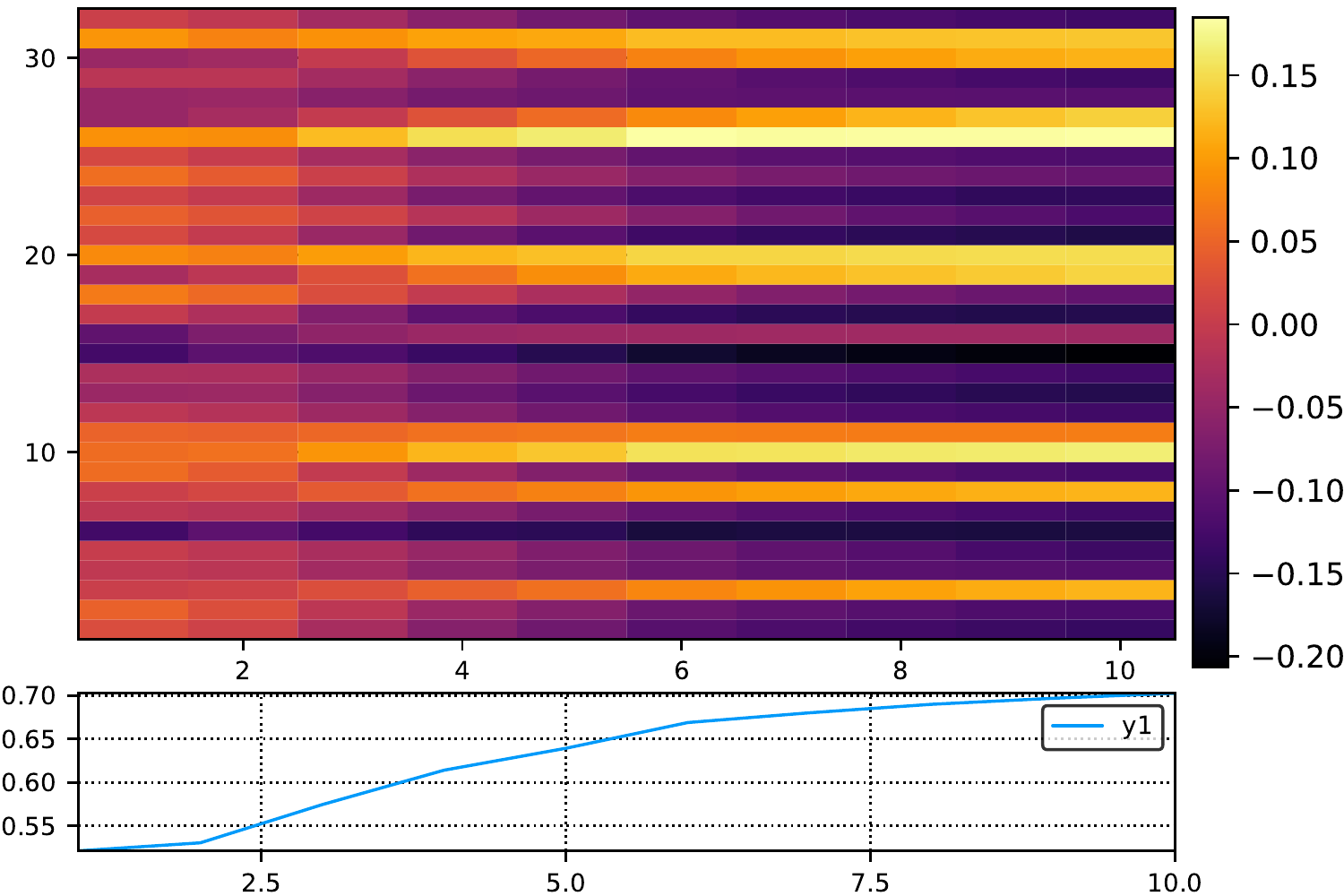} & \includegraphics[width=0.4\textwidth]{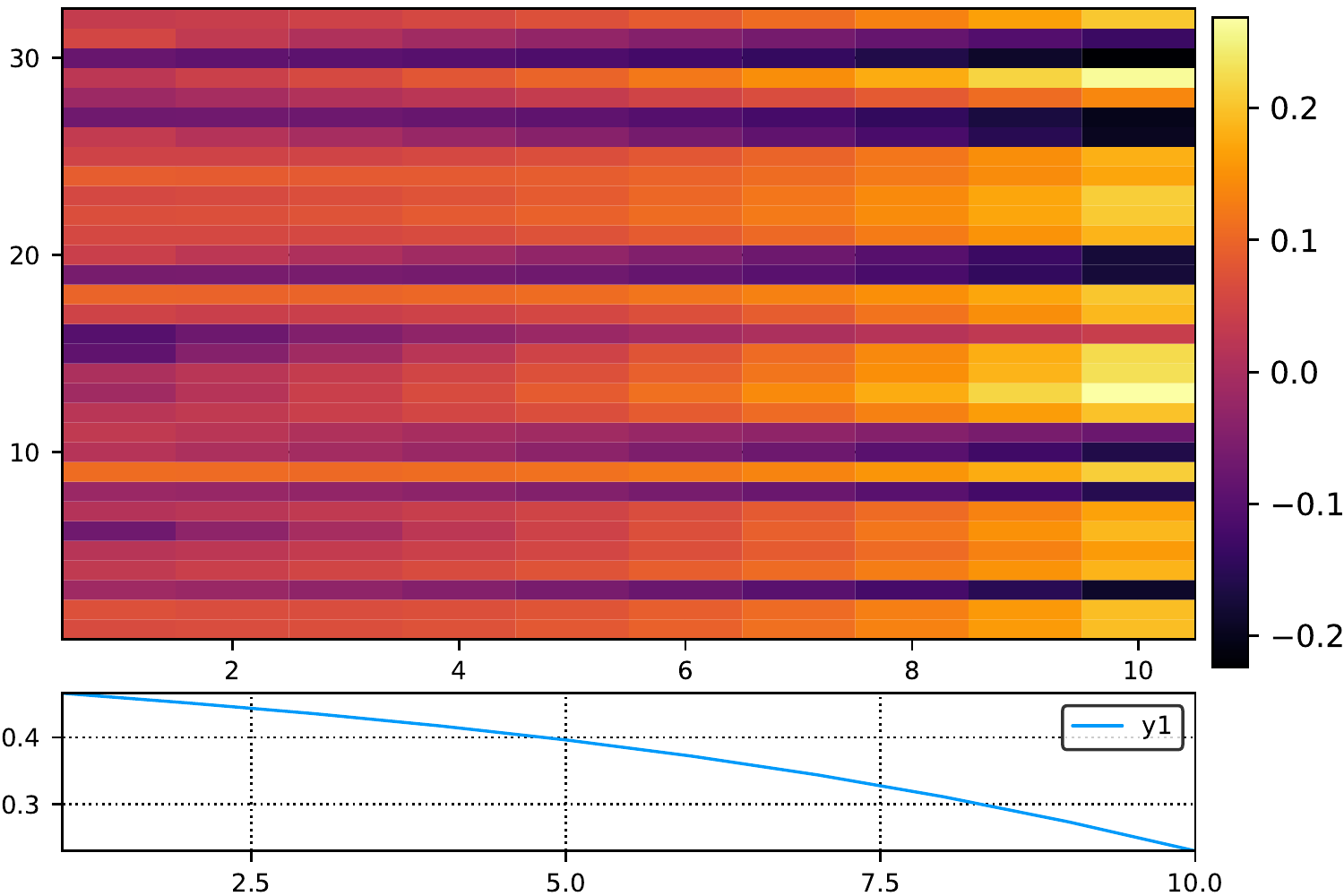}\tabularnewline
(a) Worsening progression ($P=0.70$) & (b) Improving progression ($P=0.23$)\tabularnewline
\end{tabular}
\par\end{centering}
\centering{}\caption{Illness state progression over time, measured as $\protect\hb_{t}$
for the last 10 visits. \textbf{Left figure:} a high-risk case with
70\% chance of readmission at time T=10. \textbf{Right figure}: a
low-risk case with 23\% chance at the end of the sequence. Best viewed
in color.\label{fig:hidden-states}}
\end{figure*}

The progression of illness states and probability of readmission over
time is visualized in Fig.~\ref{fig:hidden-states} for two typical
patients. The high-risk case is shown in Fig.~\ref{fig:hidden-states}(a)
\textendash{} it seems that the illness gets worse over time. In contrast,
the low-risk case is depicted in Fig.~\ref{fig:hidden-states}(b),
where the illness is rather stable over time.

\paragraph{Prediction accuracy}

Table~\ref{tab:AUC-all-methods} reports the Area Under the ROC Curve
(AUC) for all methods in predicting unplanned readmission. The proposed
methods shows a competitive performance against the baselines. The
Multi-Disease\textendash Multi-Treatment method shows better prediction
rate in the mental health data while the Multi-Disease\textendash Multi-Treatment
with Progression method information seems better in the diabetes data.
It suggests that a proper modeling of care over time is needed, not
only for understanding the underlying processes, but also to achieve
a competitive predictive performance.

\begin{table}
\begin{centering}
\begin{tabular}{lcc}
\hline
\emph{Method} & \emph{Diabetes} & \emph{Mental health}\tabularnewline
\hline
BoW+LR & 0.673 & 0.705\tabularnewline
Deepr \cite{nguyen2016deepr} & 0.680 & 0.714\tabularnewline
\hline
\textbf{MDMTP+LTSM} & \textbf{0.718} & \textbf{0.726}\tabularnewline
\textbf{MDMT+LSTM} & \textbf{0.701} & \textbf{0.730}\tabularnewline
\hline
\end{tabular}
\par\end{centering}
\caption{Area Under the ROC Curve averaged over 5 folds in predicting unplanned
readmission. BoW = bag-of-words, LR = logistic regression, MDMT =
Multi-Disease\textendash Multi-Treatment, MDMTP = Multi-Disease\textendash Multi-Treatment
with Progression information. \label{tab:AUC-all-methods}}
\end{table}

\section{Related Work}

The past few years have witnessed an intense interest in applying
recent deep learning advances to healthcare \cite{ravi2017deep}.
The most ready area is perhaps medical imaging \cite{greenspan2016guest}.
Thanks to the record-breaking successes in convolutional nets in computer
vision, we now can achieve diagnosis accuracy comparable with experts
in certain sub-areas such as skin-cancer \cite{esteva2017dermatologist}.
However, it is largely open to see if deep learning succeeds in other
areas where data are less well-structured and of lower quality such
as electronic medical records (EMR) \cite{pham2017predicting}.

Within EMRs, three set of techniques have been employed. The first
is finding distributed representation of medical objects such as diseases,
treatments and visits \cite{choi2016multi,tran2015learning}. The
techniques are not strictly deep but they offer a compelling algebraic
view of healthcare. The second group of techniques involve 1D convolutional
nets, which are designed for detecting short translation invariant
motifs over time \cite{cheng2016risk,nguyen2016deepr}. The third
group, to which this paper belongs, employs recurrent neural nets
to capture the temporal structure of care \cite{choi2016retain,pham2017predicting}.

Predictive healthcare begs the question of modeling treatment effects
\emph{over time}. This has traditionally been in the realm of randomized
controlled trials. Our work here is, on the other hand, based entirely
on observational administrative data stored in Electronic Medical
Records.

\section{Discussion}

The continuous representation of diseases make it easy to study the
disease space, that is, which diseases are related and may be interacting.
The same holds for the treatments. The interaction between diseases
and treatments can be modelled easily.

\subsection{Conclusion}

In summary, we have introduced a temporal model of care, where the
emphasis is to build a continuous representation of discrete medical
entities such as diseases, treatments and hospital visits. Representing
disease comorbidity is via simple algebraic manipulation of disease
vectors. The same holds for the care package, which is a bag of treatments
targeted at multiple diseases present in the patient. Multi-disease\textendash multi-treatment
interaction is a function of difference between the comorbidity vector
and the care package vector. A healthcare trajectory is then modelled
using a recurrent neural network known as Long Short-Term Memory,
which is capable of memorizing distant events. Importantly, the entire
system is \emph{end-to-end}: the model reads the Electronic Medical
Record and predicts future risks \emph{without} any manual feature
engineering. Inititial results on two chronic cohorts, the diabetes
and the metal health, demonstrate the usefulness of the model.

Future work will include investigation into (a) interaction models
between diseases and treatments, (b) disease progression, where we
conjecture that simple linear albegra operators such as matrix multiplications
can be used.

\paragraph{Acknowledgments}

The paper is partly supported by the Telstra-Deakin CoE in Big Data
and Machine Learning.\pagebreak{}
\bibliographystyle{plain}

\end{document}